\documentclass[lettersize,journal]{IEEEtran}
\usepackage{amsmath,amsfonts}
\usepackage{hyperref}
\usepackage{array}
\usepackage[caption=false,font=normalsize,labelfont=sf,textfont=sf]{subfig}
\usepackage{textcomp}
\usepackage{stfloats}
\usepackage{url}
\usepackage{verbatim}
\usepackage{graphicx}
\usepackage{lipsum}
\usepackage{makecell, multirow, booktabs}
\usepackage{caption}
\usepackage[space]{cite}
\usepackage[OT1]{fontenc}

\makeatletter
\newcommand{\removelatexerror}{\let\@latex@error\@gobble}
\makeatother

\usepackage[ruled,norelsize, linesnumbered]{algorithm2e}

\begin{document}

\title{MARC: Multipolicy and Risk-aware Contingency Planning
for Autonomous Driving}

\author{Tong Li$^{1}$, Lu Zhang$^{1}$, Sikang Liu${^2}$, Shaojie Shen${^1}$

\thanks{Manuscript received April 28, 2023; accepted August 21, 2023. This paper was recommended for publication by Editor Hyungpil Moon upon evaluation of the reviewers’ comments. This work was supported by the Hong Kong Ph.D. Fellowship Scheme, The Research Grants Council General Research Fund(RGC GRF) project RMGS20EG20, and the HKUST-DJI Joint Innovation Laboratory.\textit{(Corresponding author: Tong Li.)}}
\thanks{$^{1}$T. Li, L. Zhang and S. Shen are with the Department of Electronic and Computer Engineering, Hong Kong University of Science and Technology, Hong Kong (e-mail: tlibm@ust.hk; lzhangbz@ust.hk; eeshaojie@ust.hk).}
\thanks{$^{2}$S. Liu is with the DJI Technology Company, Ltd., Shenzhen, China (e-mail: sikang.liu@dji.com).}}


\maketitle

\begin{abstract}
Generating safe and non-conservative behaviors in dense, dynamic environments remains challenging for automated vehicles due to the stochastic nature of traffic participants' behaviors and their implicit interaction with the ego vehicle. This paper presents a novel planning framework, \underline{M}ultipolicy \underline{A}nd \underline{R}isk-aware \underline{C}ontingency planning (MARC), that systematically addresses these challenges by enhancing the multipolicy-based pipelines from both behavior and motion planning aspects. Specifically, MARC realizes a critical scenario set that reflects multiple possible futures conditioned on each semantic-level ego policy. Then, the generated policy-conditioned scenarios are further formulated into a tree-structured representation with a dynamic branchpoint based on the scene-level divergence. Moreover, to generate diverse driving maneuvers, we introduce risk-aware contingency planning, a bi-level optimization algorithm that simultaneously considers multiple future scenarios and user-defined risk tolerance levels. Owing to the more unified combination of behavior and motion planning layers, our framework achieves efficient decision-making and human-like driving maneuvers. Comprehensive experimental results demonstrate superior performance to other strong baselines in various environments.
\end{abstract}


\section{Introduction}
\IEEEPARstart{O}{ver} the past decade, autonomous driving has experienced remarkable progress. Nonetheless, ensuring safe and efficient operations in highly interactive environments remains a formidable challenge. The difficulties of planning under uncertainty result from imperfect observation and, more importantly, the inherently multimodal intentions of various road users that cannot be observed directly. A false estimation of the other's purpose can lead the autonomous vehicle to generate overly cautious or hazardous driving behavior, putting traffic safety at risk.

\indent The academic community has extensively explored methods to address the challenges. For interactive decision-making, existing methods~\cite{kaelbling1998planning,liu2015situation,hubmann2018automated,luo2018porca} usually utilize Markov decision process (MDP) and partially observable Markov decision process (POMDP), mathematically rigorous formulations for decision processes in stochastic environments. However, solving such problems becomes computationally intractable as the size increases. To address this issue, multipolicy-based pipelines~\cite{cunningham2015mpdm,galceran2017multipolicy,zhang2020efficient} are proposed, which prune the belief trees heavily and decompose the original problem into a limited number of closed-loop policy evaluations. These methods leverage domain-specific semantic-level policies (e.g., lane-keeping and lane-changing) to approximate the action space and realize multiple future scenarios for each policy to handle interaction uncertainties. However, these methods are typically designed to generate the \textit{best} policy over all possible future evolutions, while the subsequent trajectory generation modules only account for a single selected scenario, making the whole system hard to exploit the multi-modality of decision-making.

\begin{figure}[!t]
\centering
\includegraphics[width=0.48\textwidth]{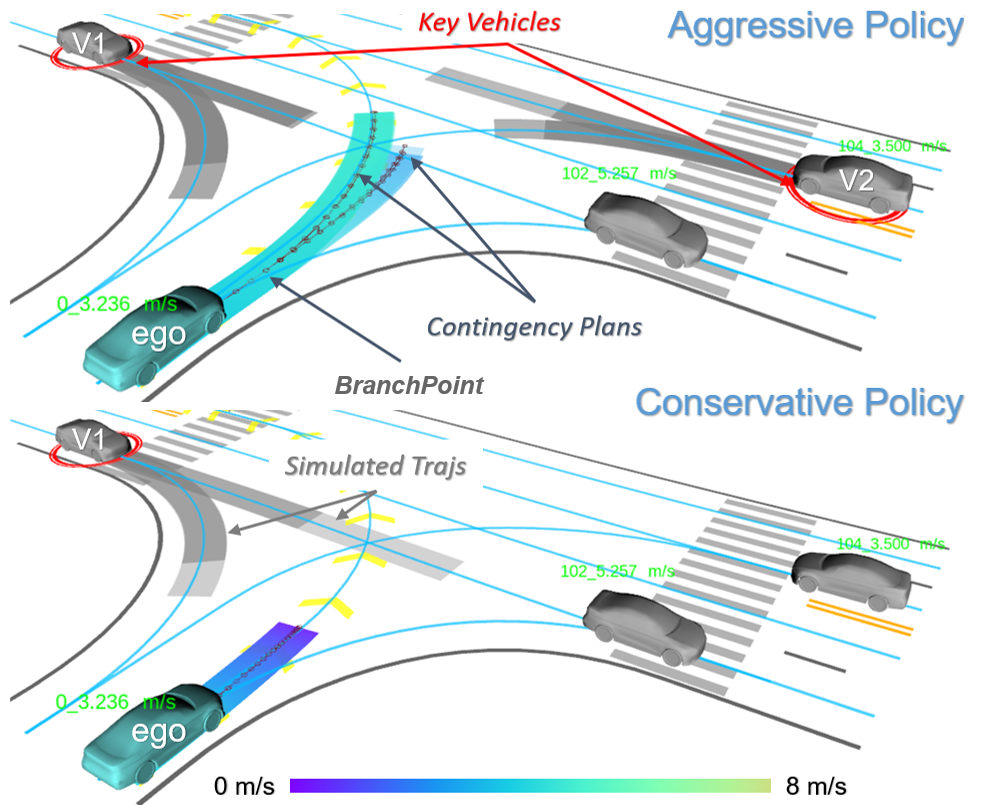}
\caption{Illustration of different policies generated by MARC. Policy-conditioned critical scenarios are rendered with different key vehicles. The darkness of trajectories indicates the associated probabilities. Scenario trees are adopted to depict the topology of rendered scenarios, and trajectory trees with risk-aware maneuvers are generated. The aggressive policy shows a siding action considering multi-modal interactions with the surrounding key vehicles, while the conservative policy converges to a smooth deceleration maneuver.}
\label{fig_1}
\vspace{-0.6cm}
\end{figure}

\indent Another way of interaction handling is generating contingency plans for multiple possible futures in the motion planning layers. Unlike the aforementioned decision-making algorithms, contingency planning~\cite{scokaert1998min, hardy2013contingency} is always formulated as a numerical optimization problem, in which the environmental uncertainties can be properly handled by optimizing a tree-structured trajectory with each branch accounting for a potential hazard. While existing methods~\cite{hardy2013contingency,chen2022interactive,da2022comprehensive} have demonstrated the ability to handle stochastic interactions with other road users, their limited ability to handle a small number of surrounding agents may pose challenges for real-world applications. Moreover, existing approaches introduce strong assumptions on the upper-stream modules, making them poorly integrated into the existing autonomous driving stack.

\indent To overcome these limitations, we proposed the MARC framework, a compact combination of behavioral and motion planning. We first generate critical scenario sets conditioned on semantic-level ego policies using forward simulations. By conducting the divergence assessment of future scenarios, we construct scenario trees with dynamic branchpoints. To diversify driving behaviors with tradeoffs between conservativeness and efficiency, we introduce \textit{risk-aware contingency planning} (RCP) by advancing contingency planning with risk measurement. The proposed RCP generates trajectory trees considering diverse risk tolerances and future scenarios based on efficient bi-level optimizations with \textit{linear programming} (LP) and the \textit{iterative linear quadratic regulator} (iLQR). Finally, the policy selection is conducted using the evaluation of both scenario trees and trajectory trees, thereby minimizing the loss of multi-modal information. We demonstrate the promising ability of MARC through comparison with a strong baseline, quantitative ablation tests, and qualitative experiments on a self-built multi-agent platform and the open-source CARLA simulator~\cite{Dosovitskiy17}. The contributions of MARC include:
\begin{itemize}
  \item MARC generates policy-conditioned critical scenarios and dynamically builds scenario trees that account for the scene-level divergence for each policy. 
  \item Risk-aware contingency planning is introduced to efficiently generate diverse maneuvers which handle multi-modal interactions under different risk tolerances.
  \item We conducted extensive quantitative experiments, comparisons with strong baselines, and qualitative experiments to showcase the capability of MARC.
\end{itemize}

\indent The remaining content is organized as follows. The related work is reviewed in Sec.~\ref{sec:related_work}, and an overview of the proposed framework is presented in Sec.~\ref{sec:sys_overview}. The methodologies are detailed in Sec.~\ref{sec:scen_tree} and Sec.~\ref{sec:risk_contin_plan}. Implementation details and experimental results are provided in Sec.~\ref{sec:imple_detail} and Sec.~\ref{sec:exp_result}. Finally, the article is concluded in Sec.~\ref{sec:concl}.

\vspace{-0.2cm}

\section{Related Work}\label{sec:related_work}
\subsection{Decision-making under uncertainty}
Extensive literature exists on addressing interactive planning under uncertainty in dynamic scenarios. POMDP provides a theoretically sound framework for handling multiple potential futures in a general form. However, due to the \textit{curse of dimensionality}, solving POMDPs in real-time is infeasible as the driving context becomes more complex, even with advanced POMDP solvers~\cite{ye2017despot,kurniawati2016online} and over-tailored problem modeling~\cite{liu2015situation,hubmann2018automated,luo2018porca}. To further enable real-time decision-making, multipolicy-based methods are proposed to simplify the POMDP process using domain-specific knowledge of autonomous driving. Particularly, multipolicy-based methods prune the belief tree of the POMDP process using semantic-level policies which approximate the original action space and retain the interaction among agents using closed-loop forward simulations. MPDM~\cite{cunningham2015mpdm, galceran2017multipolicy}, a representative of multipolicy-based methods, shows a promising ability for interactive planning. Following the spirit of MPDM, EUDM~\cite{zhang2020efficient} considers action and intention branching during the planning horizon to enable more flexible decision-making in dynamic scenarios. The EPSILON framework extends forward simulation using advanced driver models to generate human-like behavior with improved safety, and is validated through real-world autonomous driving experiments~\cite{ding2021epsilon}. TPP~\cite{Chen2023TreestructuredPP} utilizes deep learning prediction models to generate MDP with better scalability. Nevertheless, drawbacks exist. For instance, the multipolicy-based pipelines only consider policies that handle specific interactions in a single evolution and output the optimal policy over all scenarios, resulting in the lack of guarantee of policy consistency. Furthermore, existing methods only adopt single trajectory generation methods causing unavoidable loss of multi-modality information.

\vspace{-0.2cm}

\subsection{Motion Planning with Contingency}
Motion planning is another extensively studied area in autonomous driving~\cite{gonzalez2015review}. Techniques of motion planning can be categorized into search-based methods~\cite{rufli2010design,mcnaughton2011motion,ma2015efficient}, optimization-based methods~\cite{xu2012real,ziegler2014trajectory, chen2017constrained}, and a combinations of these two~\cite{gu2015tunable,fan2018baidu}. Being an optimization-based method, contingency planning generates deterministic actions for multiple future scenarios. In particular, it outputs trajectory trees with one shared segment and multiple branches to handle different problem setups independently. Using the model predictive control (MPC) to solve the branching trajectory tree with one shared state to track a desired path while considering the potential dangers, contingency MPC~\cite{alsterda2019contingency} shows the ability to counteract potential threats in static environments. Contingency planning can also generate deterministic actions, accounting for the motion uncertainty of other agents when combined with probabilistic obstacle predictions~\cite{hardy2013contingency}. Branch MPC~\cite{chen2022interactive,wang2022interaction} optimizes trajectory trees with multiple branchpoints based on scenario trees, taking into account the potential future branching of other agents' actions. Adapting similar ideas, reactive iLQR~\cite{da2022comprehensive} generates flexible yet safe maneuvers accounting for future environments without rigorously
avoiding the whole reachable regions of other agents. However, existing methods have limitations in generating effective interactions in complex circumstances with low computational costs, primarily due to the lack of guidance from the decision-making layer. Additionally, these methods' fixed or predefined scenario trees do not generalize well to real-world applications.

\indent We address the above issues by systematically combining multipolicy decision-making and contingency planning. Our framework generates a scenario tree that extracts the critical driving context for each semantic-level policy. Based on the scenario tree, we achieve efficient risk-aware maneuver generations accounting for critical scenarios with user-defined risk tolerances leveraging contingency planning under risk measurement. The compact design of our method enables consistent policy generations under interaction uncertainty while minimizing information loss between layers.

\begin{figure}[tbhp]
\centering
\includegraphics[width=0.475\textwidth]{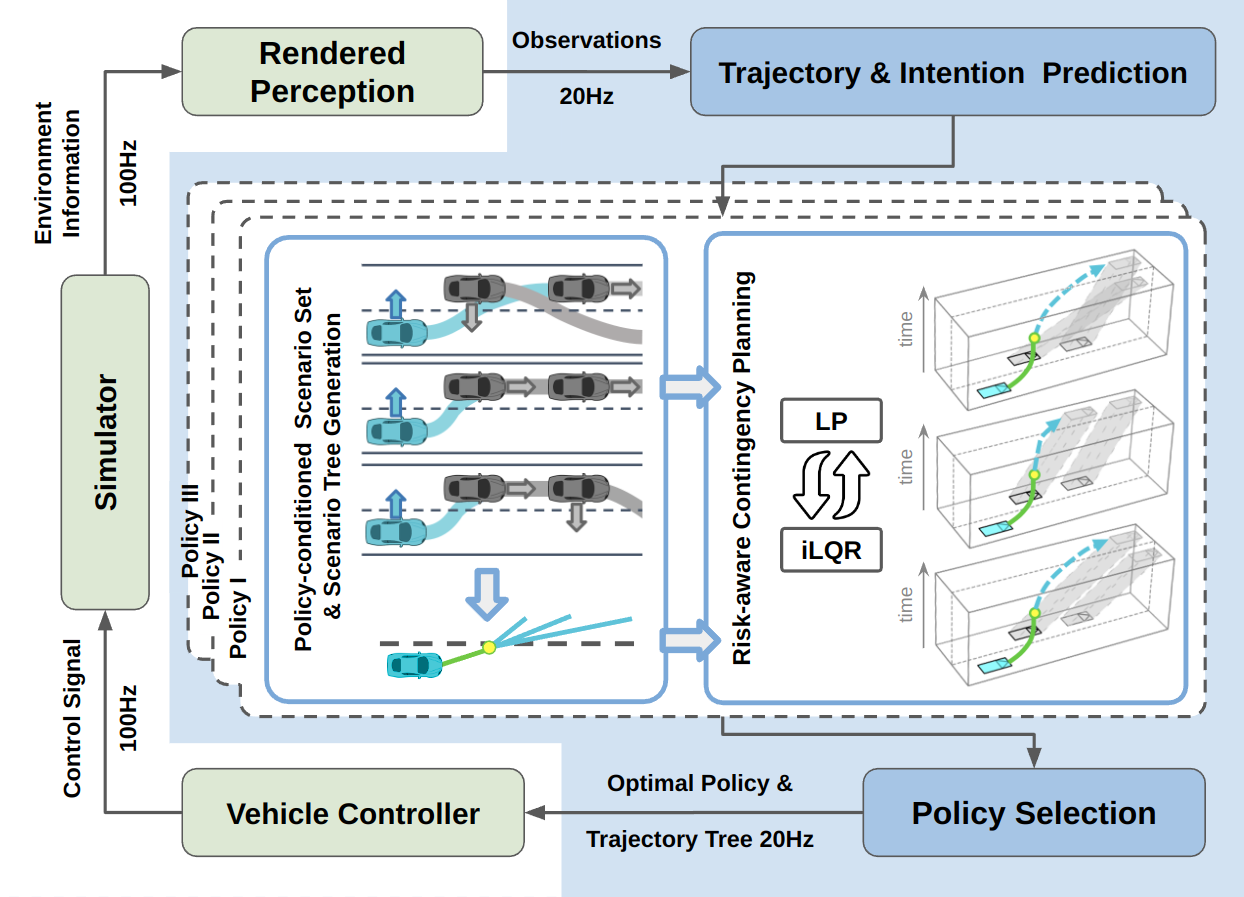}
\caption{Illustration of the MARC framework (marked in blue) and its relationship with other components. The ego vehicle is in \textit{cyan}, and surrounding vehicles are in \textit{grey}. In the scenario set, the \textit{cyan} arrows indicate the ego policies while the \textit{grey} arrows indicate the intentions of surrounding vehicles. The solid lines represent the simulated trajectories. In the scenario tree and trajectory tree, the branchpoint is in \textit{yellow}, the shared part of the trajectories is in \textit{green}, and the contingency plans are in \textit{cyan}.}
\label{fig:framework}
\vspace{-0.6cm}
\end{figure}
\vspace{-0.2cm}

\section{System Overview}\label{sec:sys_overview}
As shown in \autoref{fig:framework}, we present a complete planning pipeline that incorporates perception data from the simulator and generates optimal policy with trajectory trees for the vehicle controller. The vehicle controller evaluates certain target states on the trajectory trees and computes the corresponding control signals. These control signals, along with those from other agents, are fed into the simulator, which synchronously propagates the states according to the underlying physical models.

\indent MARC first predicts nearby vehicles' future trajectories and semantic-level intentions based on historical observations and environmental context. The critical scenario sets conditioned on the ego policies are then elected utilizing multi-agent forward simulation. The associated scenario trees with dynamic branch points are constructed utilizing scene-level divergence assessment. Taking the scenario trees, \textit{risk-aware contingency planning} (RCP) generates optimal maneuvers with the user-defined risk tolerances. Specifically, RCP solves a bi-level optimization consisting of LP for risk measurement and iLQR for contingency planning. Leveraging the multi-modality in both behavioral and motion planning layers, MARC generates actions that consider both risk and multiple future evolutions.

\section{Policy-conditioned Scenario Tree}\label{sec:scen_tree}
Using tree structures to represent scenarios provides a clear topology of future evolution. As elaborated in Sec.~\ref{sec:related_work}, fixed or predefined scenario trees have poor generalization. In contrast, our approach involves constructing dynamic scenario trees that adapt to the change of interactions. The process is conducted in a two-step manner: generating policy-conditioned critical scenario sets via closed-loop forward simulations and building tree representations with scene-level divergence assessments.

\indent We render scenarios based on the semantic-level ego policies and nearby agents' intention predictions. The computational complexity of the process is bounded by only considering policy-conditioned critical scenarios that might lead to potential risks. This pruning process is achieved by key vehicle selections with safety evaluations of non-cooperative behaviors for each intention combination. Following previous methods~\cite{galceran2017multipolicy,ding2021epsilon}, scenarios are rendered through closed-loop forward simulations using the selected vehicles with predicted intentions. Deterministic actions generated by the original forward simulation assume all agents follow basic traffic rules and interact according to pre-defined models. However, hand-crafted models may diverge from reality, leading to unrealistic simulations in extreme circumstances. We adopt the idea of \textit{forward reachable sets}(FRSs)~\cite{Kousik20, Vaskov}, which are used to provide safety guarantees for ego trajectory planning and extend them to capture the surrounding agents' potential tendencies in the simulation horizon. Therefore, the trajectory can generate open-loop responses swerving away from potential collision areas to ensure safety when closed-loop reactions are infeasible. To further ensure safety, we introduce fallback policies that take all FRSs as input and generate passive maneuvers such as pull-over and emergent braking. Hence, we ensure the completeness in the rendered scenario sets covering the agents' multi-modal behaviors conditioned on ego policies.

\indent A typical scenario tree starts from the current state and branches into different evolution at the same timestamp, representing the \textit{shared} and \textit{independent} solution spaces of the rendered scenarios, respectively. Here, the process of building the scenario tree is primarily concerned with determining the \textit{latest} timestamp at which the scenarios diverge. Delaying the branching time as much as possible offers two advantages. Firstly, the number of state variables in the trajectory optimization problem decreases as the duration of the shared part increase. More importantly, a later branching point gives the ego vehicle more reaction time to handle different potential outcomes smoothly. Since the ego trajectories are the cumulative results of the interaction with the scene context, their differences can be used to reveal the divergence of scenarios. Therefore, given the critical scenario set $\mathcal{S}$, the corresponding branch time calculation can be formally described as:
\begin{align*}
& \max \tau \in \{ 0,...,\tau_{max} \} \\
\text{s.t.} \quad & \forall \mathcal{M}(s_i, s_j, \tau) < \theta,\quad (s_i, s_j) \in \{ \mathcal{S} \times \mathcal{S} \},
\end{align*}

where $\mathcal{M}$ is a function measuring the deviations between ego states at time $\tau$ given the scenario pairs $(s_i, s_j)$ from the Cartesian product of the scenario set. The state divergence threshold is predefined as $\theta$. To preserve the multi-modality and prevent the tree structure from collapsing into a single trajectory, a maximum branching time $\tau_{max}$ is set to limit the length of the shared part. Despite its simple design, the proposed method is effective in practice. A brief depiction of the divergence assessment can be found in Fig.\autoref{fig:dive_asse}.

\section{Risk-aware Contingency Planning}\label{sec:risk_contin_plan}
\subsection{Prerequisites}
Given the tree-structure descriptions of critical scenarios, we adopt contingency planning to cope with the multi-modality. As elaborated in Sec.~\ref{sec:related_work}, contingency planning aims to obtain a trajectory tree that accounts for all possible surroundings evolutions. We define the planning problem with $K$ contingency plans as a constrained nonlinear optimization:
\begin{align*}\min_{U} & \sum_{j \in I_s}l_{j}(x_{j}, u_{j}) + \sum_{k \in K}\sum_{j \in I_k}l_{j}(x_{j}, u_{j})\\
\text{s.t.} \quad &x_{i} = f(pre(x_{i}), u_{i}), \quad \forall i \in I\backslash \{0\}\\ 
 &h_{i}(x_{i}, u_{i}) \leq 0, \quad \forall i \in I,
\end{align*}

where $I$, $I_s$, and $I_k$ are the index lists of all nodes, shared nodes, and the nodes on the $k$-th contingency plan, respectively. $X:=\{x_{i} | i \in I \}$ is the trajectory states while $U:=\{u_{i} | i \in I\backslash \{0\} \}$ is the control signals. The cost function, constraint, and state-transition function are denoted by $l(\cdot)$, $h(\cdot)$, and $f(\cdot)$, respectively. $pre(\cdot)$ returns the input node's predecessor. Fig.\autoref{fig:traj_tree} depicts a general trajectory tree.

\subsection{Bi-level Risk-aware Contingency Planning}
\indent Due to the diverse risk tolerances, human drivers tend to react differently, even given the same surrounding predictions. Therefore, risk tolerances are essential for diverse maneuver generations. We adopt \textit{conditional value-at-risk} (CVaR), which is recently introduced to contingency planning in~\cite{chen2022interactive}, as a tail risk assessment measure in our framework. Given a discrete random variable of the distinct value set $K$ with associated probabilities $\{p_{k} | k \in K \}$ and risk costs $\{\xi_{k} | k \in K \}$, the CVaR can be defined as follows~\cite{rockafellar2002deviation}:
\begin{align*}
\operatorname{CVaR}_\alpha &=\max _{Q}\sum q_k p_k \xi_k\\
\text{s.t.} \quad & 0 \leq q_k \leq (1-\alpha)^{-1}, k \in K \\
&\sum q_k p_k=1,
\end{align*}
where $Q:=\{q_{k} | k \in K \}$ is a series of variables that weigh the original risk cost under constraints. $\alpha\in \left[ 0,1 \right)$ is a hyperparameter that decides the percentage of worse cases that the risk measure should take into consideration.

\begin{figure}[!t]
\vspace{-0.6cm}
\centering
\subfloat[scene divergence]{\includegraphics[width=0.125\textwidth, keepaspectratio]{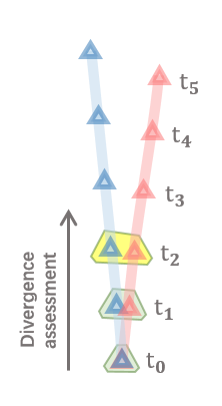}\label{fig:dive_asse}}
\hspace{\fill}
\subfloat[trajectory tree]{\includegraphics[width=0.11\textwidth, keepaspectratio]{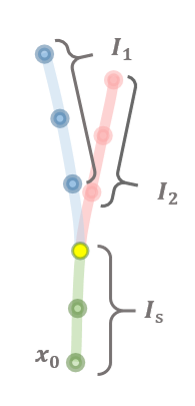}\label{fig:traj_tree}}
\hspace{\fill}
\subfloat[risk-neutral vs risk-averse]{\includegraphics[width=0.2\textwidth, keepaspectratio]
{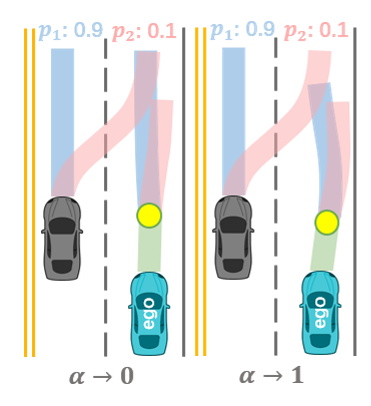}\label{fig:risk-aware}}
\caption{(a) depicts the branch time marked in \textit{yellow} selected by divergence assessment. (b) shows a trajectory tree, where $I_s$, and $I_k$ are the index lists of shared nodes, and the nodes on the $k$-th contingency plan, respectively. (c) shows maneuvers generated by RCP with different risk tolerances $\alpha$: risk-neutral (left) and risk-averse (right). The \textit{green} trajectories are the shared trajectories. The branchpoints are colored in \textit{yellow}. Contingency plans and associated predictions of agents are marked in the same colors.}
\label{fig:fig_1}
\vspace{-0.6cm}
\end{figure}

\indent To combine CVaR with contingency planning, we treat the safety part of cost functions $l^{safe}$ as the risk costs, which will be weighted by the CVaR variables $Q$ and leave the other cost functions $l^{-safe}$ unbiased. The resulting \textit{risk-aware contingency planning} (RCP) is shown as follow:
\begin{align*}
\max_{Q}\min_{U} & \sum_{j \in I_s}l_{j}(x_{j}, u_{j}) + \\
& \sum_{k\in K}\sum_{j\in I_k}\left( p_{k}q_{k}l_{j}^{safe}(x_{j}, u_{j})+l_{j}^{-safe}(x_{j}, u_{j})\right)\\
\text{s.t.} \quad &x_{i} = f(pre(x_{i}), u_{i}), \quad \forall i \in I \backslash \{0\}\\ 
 &h_{i}(x_{i}, u_{i}) \leq 0, \quad \forall i \in I \\
 \quad & 0 \leq q_k \leq \alpha^{-1}, k \in K \\
&\sum q_k p_k=1,
\end{align*}
\indent By changing $\alpha$, which indicates the risk tolerance levels of users, we can obtain diverse driving behaviors through RCP. As shown in Fig.\autoref{fig:risk-aware}, if setting $\alpha \rightarrow 0$, we can obtain risk-neutral maneuvers with the optimal variables $Q$ closed to given probabilities. Conversely, choosing $\alpha \rightarrow 1$ brings risk-averse maneuvers with increased optimal variables $Q$ weighting on the dangerous contingency plans.\\
\vspace{-0.3cm}

\indent The above problem can be solved in its dual form by introducing dual variables~\cite{chen2022interactive}. However, this method increases the problem dimension and scales poorly to the high-dimensional state-transition function. To further improve the computational efficiency, we adopt the bi-level optimization method in which the solution to the original problem is obtained by iteratively solving two subproblems using iLQR and LP.

\indent iLQR utilizes the Taylor-series expansion to obtain approximated problems consisting of quadratic cost functions and discrete system transition functions of the original problem. Given fixed risk-related variables $Q$, we can obtain the approximated subproblem for control sequence $U$ as follows:
\begin{align*}
\min_{U} & \sum_{j\in I_s}(\dfrac{1}{2} x_{j}^{T}R_{j}x_{j} + r_{j}^{T}x_{j} + \dfrac{1}{2}u_{j}^{T}S_{j}u_{j} + s_{j}^{T}u_{j}) + \nonumber \\
\sum_{k\in K}&\sum_{j\in I_k}p_{k}q_{k}(\dfrac{1}{2} x_{j}^{T}R_{j}^{safe}x_{j} + {r_{j}^{safe}}^{T}x_{j}) + \nonumber \\
\sum_{k\in K}&\sum_{j\in I_k} (\dfrac{1}{2} x_{j}^{T}R_{j}^{-safe}x_{j} + {r_{j}^{-safe}}^{T}x_{j} + \dfrac{1}{2}u_{j}^{T}S_{j}u_{j} + s_{j}^{T}u_{j}) \\ 
&\text{s.t.} \quad x_{i} = A_{i}pre(x_{i}) + B_{i}u_{i}, \quad \forall i \in I \backslash \{0\},
\end{align*}
where $A_{i}\equiv\frac{\partial f}{\partial pre(x_{i})}$ and $B_{i}\equiv\frac{\partial f}{\partial u_{i}}$. This approximated problem could be solved by iLQR in the \textit{discrete dynamic programming} style using backward and forward propagation. We refer the interested readers to~\cite{mayne1966second}.

\indent Note that for any fixed control sequence $U$, the original problem degenerates into an LP. Solving the LP finds the CVaR with the optimal variables $Q$ under an extra equality constraint. For any fixed $Q$, the original problem can be approximated by iLQR. Solving the iLQR subproblem obtains optimal controls $U$ of the trajectory tree. Since both subproblems are convex, we can derive a bi-level optimization with a convergence guarantee:
\begin{align*}
U^{k+1} &=\underset{U}{\operatorname{argmin}}\enspace \operatorname{iLQR}\left(X^{k}, U^{k}, Q^{k}\right) \\
Q^{k+1} &=\underset{Q}{\operatorname{argmax}}\enspace \operatorname{LP}\left(X^{k+1}, U^{k+1}, Q^{k}\right).
\end{align*}

\indent Owing to the low cost of solving LP and iLQR, the computational efficiency of the risk-aware contingency planning can be guaranteed. Compared with original contingency planning, RCP generates trajectories that are optimal in multiple future scenarios under user-defined risk-averse levels, which can give rise to more human-like behaviors.

\subsection{Policy Evaluation}
\indent As stated in Sec.~\ref{sec:related_work}, existing multipolicy-based pipelines usually require a subsequent motion planner that only considers a single selected scenario, leading to potential oscillations in the final output when dealing with rapidly changing scenarios. In contrast, our framework conducts policy selection based on the evaluation of both scenario trees and trajectory trees, ensuring the consistency of the resulting policies.

\indent The flow of the MARC framework is detailed in \autoref{alg:1}. Policy-conditioned scenarios are rendered under each policy based on the intention predictions (Line 7 to Line 10). The scenario trees and trajectory trees are obtained consecutively from the critical scenario sets (Line 11 to Line 12). Evaluations are conducted based on both scenario trees and trajectory trees (Line 14). Most parts of our framework (Line 5 to Line 15) can be run in parallel for real-time performance.
\begin{figure}[!t]
\removelatexerror
\SetKwInOut{Input}{Inputs}
\SetKwInOut{Output}{Outputs}
\SetKwComment{Comment}{//}{}
\SetKwComment{SComment}{/* }{}
\begin{algorithm}[H]
\LinesNumbered
\caption{Process of MARC}\label{alg:1}
\Input{Perception $Z$, Ego vehicle state $x$, Ego policy set $\Pi$, Horizon $\tau_h$, Max Branch Time $\tau_{max}$, Divergence threshold $\theta$}
\Output{Optimal policy $\pi^{*}$, Optimal trajectories $\mathcal{T}^{*}$}
$\mathcal{R} \gets\emptyset$ \Comment*[l]{ init policy reward set}
$\mathcal{T} \gets\emptyset$ \Comment*[l]{ init policy trajectory set}
$\mathcal{P}, \mathcal{I} \gets $ PredictIntentionTrajectories($Z, x$) \;
$\mathcal{I}_{c} \gets $ GetIntentionCombination($\mathcal{I}$) \;
\For{$\pi \in \Pi$}{
  \SComment{Scenario Generation */}
  $\mathcal{S_\pi} \gets\emptyset$ \Comment*[l]{ init scenario set}
  \For{$c \in C$}{
    $V_k \gets $ GetKeyVehicles($Z, c, x, \pi$) \;
    $\mathcal{S_\pi} \gets $ ForwardSimulation($Z, c, V_k, x, \pi, \tau_h$) \;
  }
  $\Psi_\pi \gets $ ScenarioTreeConstruction($\mathcal{S_\pi},\tau_{max},\theta$) \;
  $\mathcal{T}_\pi \gets $ RcpOptimization($\pi, \Psi_\pi, \tau_h$) \;
  $\mathcal{T} \gets \mathcal{T}_\pi$ \;
  $\mathcal{R} \gets $ PolicyEvaluation($\pi, \Psi_\pi, \mathcal{T}_\pi$) \;
}
$\pi^*, \mathcal{T}^* \gets $ PolicySelection($\mathcal{R}, \mathcal{T}$) \;
\end{algorithm}
\vspace{-0.6cm}
\end{figure}

\section{Implementation Details}\label{sec:imple_detail}
\subsection{Trajectory and Intention Prediction}

We adopt a neural network-based motion predictor based on~\cite{zhao2019multi} to provide scene-centric multi-modal trajectory prediction for all nearby agents. The network is trained on the Argoverse dataset~\cite{chang2019argoverse} and deployed in C++ via LibTorch\footnote{The C++ distribution of PyTorch. See https://pytorch.org for details.}. After obtaining the predicted trajectories and their probability scores, the next step is to transform them into semantic-level intentions. In this work, the longitudinal intentions of other agents are defined as a set of semantic actions (i.e., maintaining speed, acceleration, and deceleration), while the lateral intentions are represented as the reference lane sequences. To extract the lateral intentions, a simple voting method is used based on the predicted trajectories' positional distribution. In particular, we begin by extracting candidate reference lanes for the target agent and then associate each predicted trajectory with these candidate lanes to evaluate the likelihood of each lane. Once the lateral intention is obtained, we use a similar method to estimate the longitudinal intentions and associated probabilities. We note that it is possible to predict both future trajectories and intentions using a single network, which we leave as future work.

\subsection{Ego Policy Design}
Considering achieving diverse driving behaviors, we define semantic-level policies for the ego vehicle, which can be categorized as longitudinal and lateral policies. Both longitudinal and lateral policies encompass basic semantic-level policies and scene-related policies. For example, in urban driving, the basic longitudinal policies could be maintaining speed, decelerating, and accelerating. Scene-related policies could be stopping at a specific position (i.e., stop line and conflict zone), decelerating to yield, and accelerating to the speed limit. Similarly, the lateral policies can be defined as basic policies such as lane keeping, lane change to the target lane sequence, and scene-related policies such as bypass and in-lane siding.

\subsection{Forward Simulation with FRSs}
Existing work usually obtains reachable sets with model-based propagation or over-approximations~\cite{kousik2017safe, hwang2003applications}. In this work, we adopt a simple yet effective method to generate FRSs. Specifically, we directly compute the spatial-temporal occupancy of multi-modal predictions and collect the occupancy union according to the timestamp to form the FRSs. Since FRS is essentially a spatial-temporal occupation description without probabilistic features and the trajectory predictions have implicitly encoded the estimation of the hidden state as well as the transition model of the surrounding agents, our method can approximate these occupations with little computational cost while preserving multi-modality.

\indent The forward simulation takes two steps. We first conduct the deterministic closed-loop forward simulation similar to~\cite{zhang2020efficient} with safety assessment
nts on the generated trajectories. The failure of the assessments will trigger the second simulation where we replace those failed policies with our fallback policies. By running the second forward simulation, we can obtain the final trajectories with closed-loop reactions to nearby agents and open-loop reactions to FRSs.

\subsection{iLQR Design}
We adopt the kinematic bicycle model as the state-transition function $f(\cdot)$ in iLQR and design the cost function of $i$-th node $l_i$ as the sum of safety cost~$l_{i}^{safe}$, target cost~$l_{i}^{tar}$, kinematic cost~$l_{i}^{kin}$ and comfort cost~$l_{i}^{comf}$:
\begin{gather*}
l_i = l_{i}^{safe} + l_{i}^{tar} + l_{i}^{kin} + l_{i}^{comf}.
\end{gather*}
\indent Similar to \cite{chen2017constrained}, environmental constraints can be modeled utilizing vectorized representations (such as polygons for drivable areas and polylines for reference lines). We introduce a distance function $\mathcal{D}(\cdot)$ that returns the Euclidean distance between the ego vehicle and the environmental constraint. To ensure the second-order continuity, we further define a smooth function with parameter $\beta$ as follows:
\begin{gather*}
\mathcal{S}_{\beta}(\Gamma) = \sum_{i=1}^{\Omega} \dfrac{exp(\beta \gamma_{i})\gamma_{i}}{\sum_{i=1}^{\Omega}exp(\beta \gamma_{i})}, \Gamma = \{\gamma_{1}, \gamma_{2}, ..., \gamma_{\Omega}\}^{T}.
\end{gather*}

\indent With a slight abuse of notation, we denote $\mathcal{G}(\cdot)$ as the user-designed penalty function. The safety cost penalizes the signed distances to multiple environmental constraints:
\begin{gather*}
\small l_{i}^{safe} = \overbrace{\mathcal{G}_{dri}\Big( \big(1 - \mathbf{1}_{P_{d}}\big)\mathcal{S}_{\beta < 0} \big( \mathcal{D}(P_{d})\big)\Big)}^{\text{drivable area}} + \nonumber \\
\small \overbrace{\sum_{j \in n_{p}} \mathcal{G}_{bb}\big(\mathbf{1}_{P_{b_{j}}} \mathcal{S}_{\beta < 0}(\mathcal{D}(P_{b_{j}}))\big)}^{\text{bounding box}} + \overbrace{\sum_{j\in n_{p}} \mathcal{G}_{rs}\big(\mathbf{1}_{P_{r_{j}}} \mathcal{S}_{\beta < 0} ( \mathcal{D}(P_{r_{j}}))\big)}^{\text{reachable set}},
\end{gather*}
where $P_{d}$ defines a drivable area that is feasible and collision-free to static obstacles, $P_{b_{j}}$ is the $i$-th inflated bounding boxes of $j$-th surrounding agent, $P_{r_{j}}$ is the FRS of $j$-th surrounding agent and $\mathbf{1}(\cdot)$ denotes the enclosed indicator function. Similar formulations can be found in the domain of backup policy control barrier function quadratic programming~\cite{backupCBF}.

\indent The target term varies depending on the policies and the scenarios. For instance, the reference can be a centerline that serves as a guide for lane change policies in highway scenarios, while alternatively representing a desired area for overtaking at an intersection. For simplicity, we divide this term into two parts: reference cost and desired area cost:
\begin{align*}
l_{i}^{tar} &= \mathcal{G}_{ref}\big(\mathcal{S}_{\beta < 0} \big( \mathcal{D}(L_{ref})\big) \big) + \mathcal{G}_{des}\big(\mathcal{S}_{\beta < 0} \big( \mathcal{D}(P_{des})\big) \big),
\end{align*}
where~$L_{ref}$ is the reference polyline of target lane sequence, $P_{des}$ is the target area defined by polygon.

\indent The kinematic constraints are introduced by penalizing the state and control variables that exceed the limits:
\begin{align*}
l_{i}^{kin} &= \mathcal{G}_{xub}\big(max(x_{t} - x_{ub}, 0)\big) + \mathcal{G}_{xlb}\big(max(x_{lb} - x_{t}, 0)\big) \nonumber \\
&+ \mathcal{G}_{uub}\big(max(u_{t} - u_{ub}, 0)\big) + \mathcal{G}_{ulb}\big(max(u_{lb} - u_{t}, 0)\big),
\end{align*}
where $x_{ub}$, $x_{lb}$ are the upper and lower bound of the state variables, while $u_{ub}$, $u_{lb}$ are the upper and lower bound of the control inputs, $max(\cdot)$ is the element-wise max function.

\indent The final term of the objective function is designed to minimize both the longitudinal discomfort (i.e., acceleration and jerk) and the lateral discomfort (i.e., lateral acceleration and steering rate), thereby ensuring the smoothness of the generated trajectories:
\begin{gather*}
l_{i}^{comf} = \mathcal{G}_{lon}(x_{t}) + \mathcal{G}_{lat}(x_{t}).
\end{gather*}

\subsection{Policy Evaluation}
In this work, the total policy reward is evaluated based on the scenario contexts and risk-aware contingency plans. Specifically, the reward of policy is defined as the negative weighted sum of several components:
\begin{gather*}
R = -(\lambda_1 F_s + \lambda_2 F_e + \lambda_3 F_n + \lambda_4 F_r + \lambda_5 F_u),
\end{gather*}
where the safety cost~$F_s$ is measured by the distance between trajectories of ego and other vehicles. The efficiency cost~$F_e$ is computed from the gap between the average velocity of contingency plans and desired velocity. The navigation cost~$F_n$ is obtained by the difference between the routing and target lane sequence. The risk cost~$F_r$ is obtained from the cost of RCP, and the uncertainty cost~$F_u$ is measured by the branch time and the divergence between different contingency plans. 

\begin{figure}[!bth]
\vspace{-0.2cm}
\centering
\subfloat[cut-in]{\includegraphics[width=0.1325\textwidth, keepaspectratio]{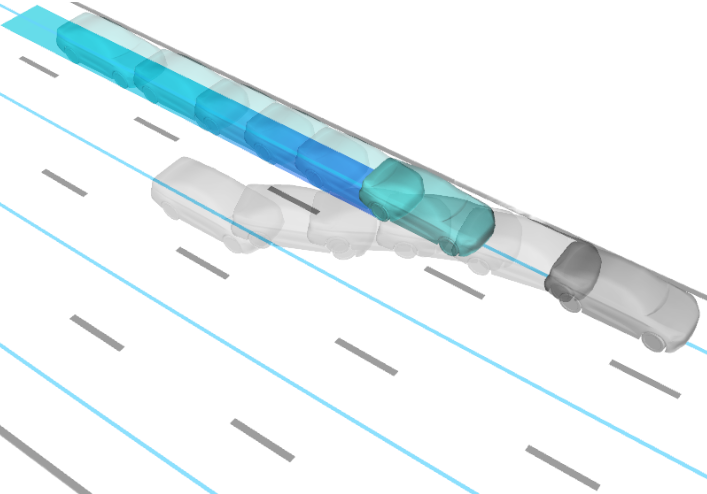}\label{fig:cut_in}}
\hspace{\fill}
\subfloat[noise injection]{\includegraphics[width=0.11\textwidth, keepaspectratio]{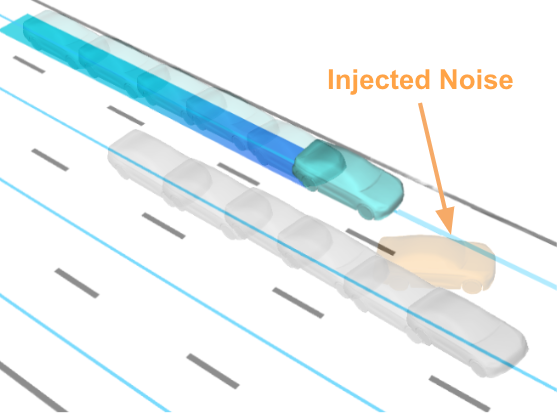}\label{fig:noise_inject}}
\hspace{\fill}
\subfloat[dynamic traffic]{\includegraphics[width=0.2\textwidth, keepaspectratio]
{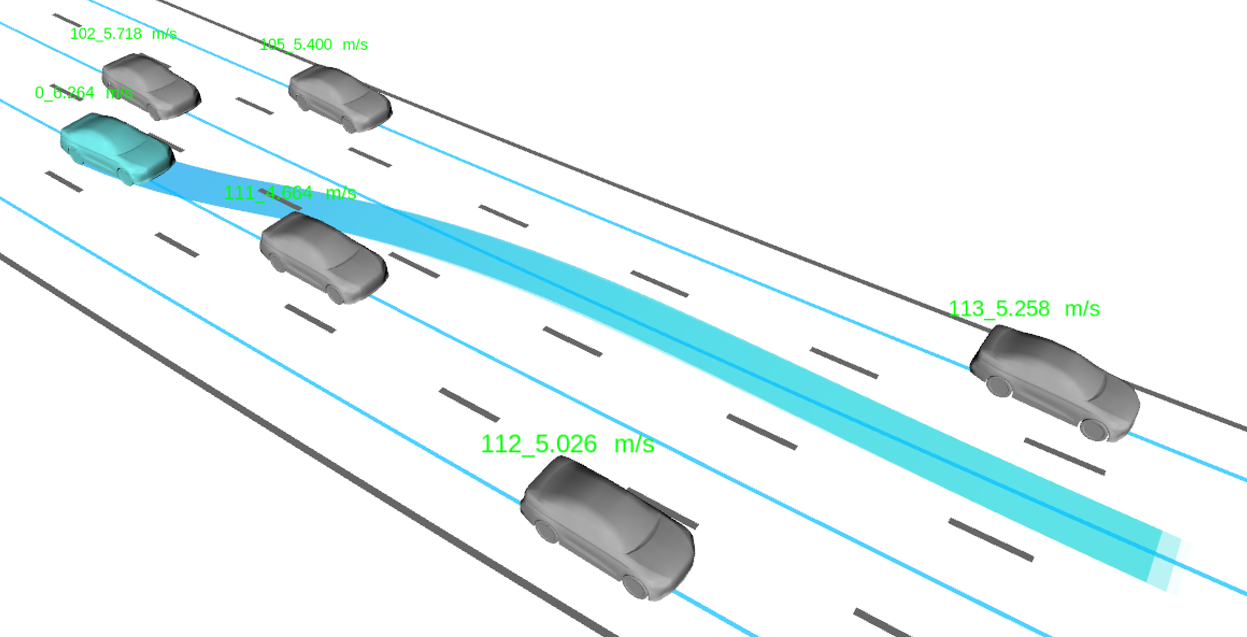}\label{fig:interactive}}
\caption{Illustration of the scenario setups. (a) shows the snapshots of the cut-in scenario. The ego vehicle is expected to yield to the non-cooperative vehicle. (b) illustrates the noise injection (marked in \textit{yellow}) on the predictions (c) is a typical dynamic traffic scenario in which the ego vehicle has to interact with other agents under uncertainty.}
\label{fig:setup}
\end{figure}

\section{Experimental Results}\label{sec:exp_result}
\subsection{Simulation Platform and Environment}
The experiments are conducted on a self-built multi-agent simulation platform elaborated in Sec.~\ref{sec:sys_overview} and CARLA. The proposed MARC framework is implemented in C++11 with customized solvers. All experiments are run on a desktop computer equipped with Intel i9-12900K CPU and 32GB RAM with a stable running rate of 20Hz.

\subsection{Quantitative Results}
We conduct comparisons with the existing strong baseline and ablation tests on our self-built platform to evaluate the performance of the proposed method quantitatively. 
\begin{figure}[t]
    \centering
    \includegraphics[width=0.48\textwidth]{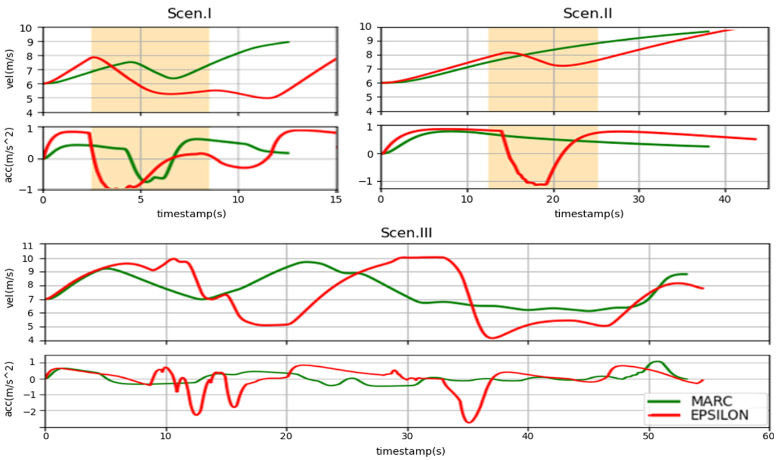}
     \caption{Dynamic profiles of the comparisons with EPSILON. In Scen.I, MARC captures the cut-in intention and generates smoother deceleration during the cut-in period marked in \textit{yellow}. In Scen.II, the additional noise is added to the predictions from 2.5s to 6s marked in \textit{yellow}. MARC handles the disturbances while EPSILON switches the policy which results in uncomfortable deceleration. In Scen.III, MARC outperforms EPSILON with better efficiency and riding comfort.}
     \label{fig:quantitative}
     \vspace{-0.2cm}
\end{figure}

\subsubsection{Comparison with EPSILON}
We conduct qualitative comparisons with the existing strong baseline EPSILON in the three scenarios. Scen.I is a non-cooperative cut-in scenario aiming to test the ability to handle extreme cases. Scen.II is a noise injection test where the adjacent vehicle remains lane-keeping during the test while noise is injected into the prediction module of the ego vehicle, increasing the lane-changing prediction, as shown in Fig.\autoref{fig:noise_inject}. This setup simulates the case where predictions deviate from the ground truth, which is common in real-world applications. Scen.III is a highway scenario with multiple autonomous agents, which is set up to assess the planning ability under intensive uncertain interactions. As shown in Fig.\autoref{fig:interactive}, the ego vehicle is assigned a global route and a speed limit to navigate through the traffic in which agents will conduct various behaviors. For fair comparisons, we use the same prediction module for both methods. To show that the policies of MARC are more robust under uncertain interactions and bring fewer unexpected policy switches, which results in better efficiency and comfort, we use total time and average speed during the test for efficiency measurement and use the root mean squared acceleration and maximum absolute acceleration for comfort assessment~\cite{failsafe,inpass}. As shown in ~\autoref{table_1}, our method outperforms the baseline with improvements in all metrics. The dynamic profiles are shown in ~\autoref{fig:quantitative} to show side-by-side comparisons in three scenarios. MARC, being more cautious when facing multiple predictions, accelerates smoother and decelerates in time in Scen.I. In Scen.II, MARC executes more consistent policies under noisy predictions, whereas EPSILON switches its policies resulting in sudden decelerations. In Scen.III, MARC handles the uncertain interactions better than EPSILON, with fewer unnecessary decelerations and a higher average speed, improving comfort and efficiency.

\begin{figure*}[t]
    \centering
    \includegraphics[width=\textwidth]{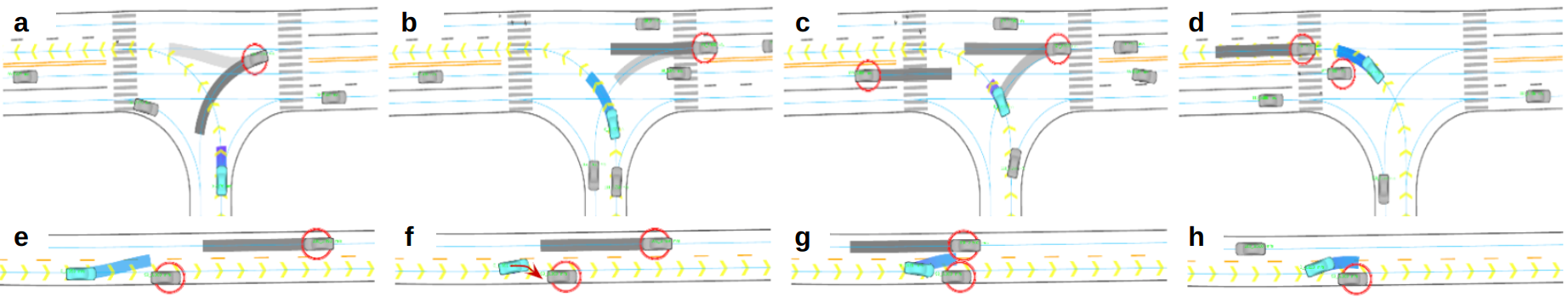}
    \caption{Key frames of the qualitative experiments. The global route is marked by \textit{yellow arrows}, key vehicles are marked by \textit{red circles}, and the simulated trajectories are marked by \textit{grey bands} with darkness representing probabilities. (a-d) Unprotected left turn: ego vehicle first yields to the aggressive left-turning vehicle~(a), slowly moves forward to stop the oncoming vehicle on the left meanwhile yields to the second vehicle on the right~(b-c), and finally pass around the stopped vehicle to exit the intersection~(d). (e-h) Bypass with oncoming traffic: ego vehicle first swerves to the left, trying to bypass the parked vehicle~(e), once anticipating the oncoming vehicle is not likely to yield, ego vehicle aborts bypassing and turn into the original lane for collision avoidance~(f), and it quickly recovers the bypass policy as soon as the spatiotemporal constraints are met~(g-h).}
    \label{fig:exp2}
    \vspace{-0.3cm}
\end{figure*}

\begin{table}
\caption{Comparison with EPSILON}
\label{table_1}
\centering
\resizebox{\columnwidth}{!}{\begin{tabular}{l|l|cccc} 
\toprule
\multicolumn{2}{c}{\textbf{Methods}} & \makecell{\textbf{Time}\\$(s)$} & \makecell{\textbf{Avg Spd}\\$(m/s)$} & \makecell{\textbf{RMS Acc} \\$(m/s^2)$} & \makecell{\textbf{Max Abs Acc}\\$(m/s^2)$} \\ 
\midrule
\multirow{2}{*}{Scen. I} & EPSILON & 19.90 & 6.87 & 0.63 & 1.03 \\ 
~ & MARC &  \textbf{12.56} & \textbf{7.26} & \textbf{0.45} & \textbf{0.75} \\ 
\hline
\multirow{2}{*}{Scen. II} & EPSILON & 8.70 & 7.90 & 0.72 & 1.13 \\ 
~ & MARC &  \textbf{7.60} & \textbf{8.03} & \textbf{0.51} & \textbf{0.78} \\
\hline
\multirow{2}{*}{Scen. III} & EPSILON & 54.46 & 7.41 & 0.70 & 2.78 \\ 
~ & MARC & \textbf{53.13} & \textbf{7.61} & \textbf{0.34} & \textbf{1.09} \\
\bottomrule
\end{tabular}}
\vspace{-0.6cm}
\end{table}

\subsubsection{Ablative tests of RCP}
We modify the cut-in scenario~\autoref{fig:cut_in} for the ablation test, in which the ego vehicle only executes the lane-keeping policy, and the neighboring vehicle conducts a non-cooperative lane-change maneuver forcing the ego vehicle to yield. A baseline planner, a fixed branching mechanism, a dynamic branching mechanism, and a dynamic branching mechanism with risk-aware optimization methods will be tested in this scenario 100 times with random initialization. We use scene-specific metrics for comfort and safety assessments: average maximum deceleration, average minimum distance with respect to the cut-in vehicle, and success rate(non-collision cases vs. total cases). Metrics such as time to collision and intervehicular time are also applicable for safety assessments~\cite{glaser2010maneuver}. As shown in \autoref{table_2}, the baseline only reacts after the most likely prediction aligns with the cut-in behaviors, resulting in the hardest deceleration, closest distance, and the lowest success rate. Branching mechanisms enables planners to response earlier, smoothing the deceleration and increasing the minimum distance. Dynamic branching outperforms fixed one owing to the better adaptation to the scenarios. Combining risk measures with dynamic branching brings a more defensive yet smoother driving style which achieves the best performance among methods, proving the effectiveness of RCP.

\begin{table}
\caption{Ablation Study Results}
\label{table_2}
\centering
\resizebox{\columnwidth}{!}{\begin{tabular}{l*{3}{c}}
\toprule
\textbf{Methods} & \makecell{\textbf{Avg Max Dec} \\ $(m/s^2)$} & \makecell{\textbf{Avg Min Dis} \\ $(m)$} & \makecell{\textbf{Suc Rate} \\ $(\%)$}\\ 
\midrule
w/o branch & 3.48 & 1.37 & 83 \\ 
Fixed branch & 2.52 & 2.53 & 93 \\ 
Dyna branch & 1.63 & 3.45 & 96 \\ 
Dyna branch + Risk & \textbf{1.14} & \textbf{4.12} & \textbf{100} \\ 
\bottomrule
\end{tabular}}
\vspace{-0.6cm}
\end{table}
\vspace{-0.2cm}
\subsection{Qualitative Results}
We conduct qualitative tests in highly interactive scenarios such as unprotected left turns and bypass with oncoming traffic on the self-built platform and scenarios with random traffic on CARLA to further verify the generalization of our method.

\subsubsection{Unprotected left turn} The ego vehicle enters a 3-way intersection with no traffic signals. Vehicles from different directions are randomly generated, and their actual intentions are unknown to the ego vehicle. As shown in \autoref{fig:exp2}~(a-d), the ego vehicle first slows down to yield to the first left-turning vehicle~(a), creeps forward to stop the vehicle on the left while giving the right of way to the second vehicle on the right~(b-c), and finally exits the intersection with a swerving maneuver to avoid the potential collision~(d). The result shows MARC's ability to handle uncertain intentions and generate reasonable behaviors during negotiations.
 
\subsubsection{Bypass with oncoming traffic} This scenario consists of a parked vehicle and a fast-moving vehicle in the opposite lane. As shown in \autoref{fig:exp2}~(e-h), the ego vehicle first swerves to the left, attempting to conduct a left bypassing maneuver on the parked vehicle~(e), once noticing that the oncoming agent is less likely to yield, it aborts bypassing policy and returns the original lane for collision avoidance~(f), and after holding to yield for a while, it switches back to the bypass policy once it anticipates the spatial-temporal constraints are satisfied~(g-h). The result reveals MARC's ability to efficient behavior generations under spatial-temporal constraints in conflict situations.  

\subsubsection{Qualitative tests on CARLA} 
\begin{figure}[t]
    \centering
    \includegraphics[width=0.48\textwidth]{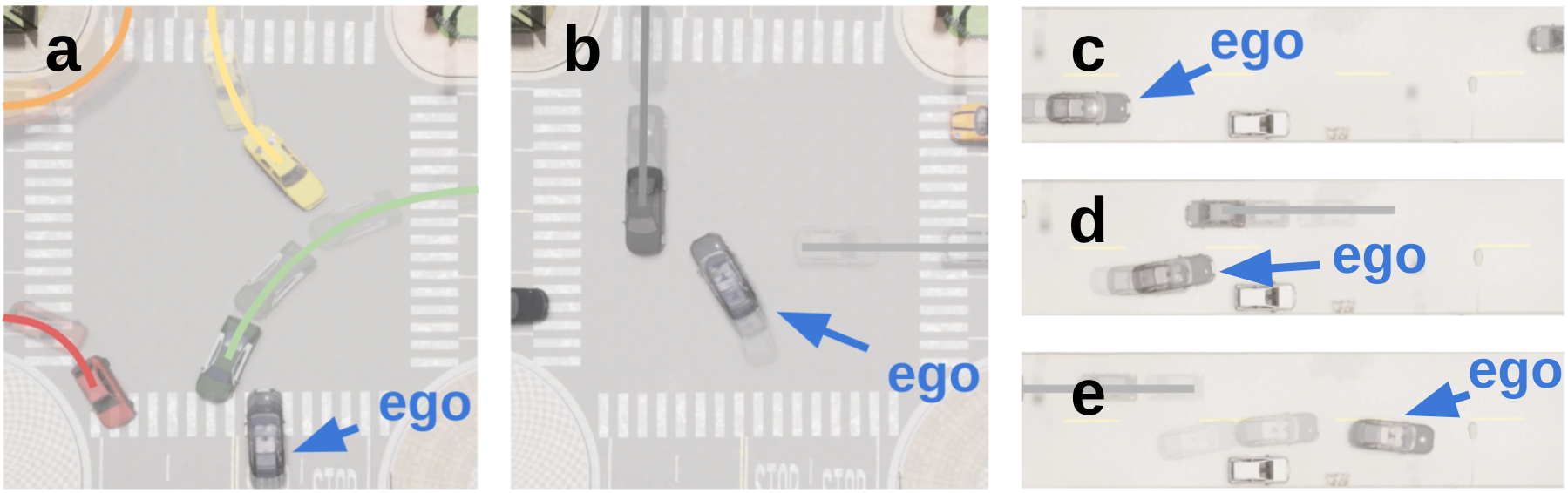}
     \caption{Keyframes of tests on CARLA. The historical positions and trajectories of surrounding agents are marked with transparent colors. (a-b)Ego vehicle achieves human-like defensive left-turning behaviors in intersection with random traffic. (c-e)Ego vehicle achieves bypassing with negotiation behaviors.}
     \label{fig:carla_exp}
     \vspace{-0.2cm}
\end{figure}

We extend our method to CARLA and conduct tests on random traffic flows. As shown in \autoref{fig:carla_exp}, MARC generates human-like driving behaviors such as defensive left-turning and bypassing under negotiation in the tests, proving its generalization in various scenarios.

\vspace{-0.2cm}
\section{Conclusion and Future Work}\label{sec:concl}
We present the MARC framework as a solution for autonomous driving in dynamic interactive environments. Our framework systematically unified two innovative techniques, policy-conditioned scenario tree and risk-aware contingency planning, to generate safe and interactive driving behaviors considering risk tolerance and uncertainty. Comparisons with a strong baseline and comprehensive experiments in challenging scenarios proved the effectiveness of our design. In the future, we will conduct real-world field tests for MARC.

\vspace{-0.2cm}

\bibliographystyle{IEEEtran}
\bibliography{ref}

\begin{thebibliography}{10}
\providecommand{\url}[1]{#1}
\csname url@samestyle\endcsname
\providecommand{\newblock}{\relax}
\providecommand{\bibinfo}[2]{#2}
\providecommand{\BIBentrySTDinterwordspacing}{\spaceskip=0pt\relax}
\providecommand{\BIBentryALTinterwordstretchfactor}{4}
\providecommand{\BIBentryALTinterwordspacing}{\spaceskip=\fontdimen2\font plus
\BIBentryALTinterwordstretchfactor\fontdimen3\font minus
  \fontdimen4\font\relax}
\providecommand{\BIBforeignlanguage}[2]{{%
\expandafter\ifx\csname l@#1\endcsname\relax
\typeout{** WARNING: IEEEtran.bst: No hyphenation pattern has been}%
\typeout{** loaded for the language `#1'. Using the pattern for}%
\typeout{** the default language instead.}%
\else
\language=\csname l@#1\endcsname
\fi
#2}}
\providecommand{\BIBdecl}{\relax}
\BIBdecl

\bibitem{kaelbling1998planning}
L.~P. Kaelbling, M.~L. Littman, and A.~R. Cassandra, ``Planning and acting in
  partially observable stochastic domains,'' \emph{Art. Intel.}, vol. 101, no.
  1-2, pp. 99--134, 1998.

\bibitem{liu2015situation}
W.~Liu, S.-W. Kim, S.~Pendleton, and M.~H. Ang, ``Situation-aware decision
  making for autonomous driving on urban road using online {POMDP},'' in
  \emph{IV. IEEE}, 2015, pp. 1126--1133.

\bibitem{hubmann2018automated}
C.~Hubmann, J.~Schulz, M.~Becker, D.~Althoff, and C.~Stiller, ``Automated
  driving in uncertain environments: Planning with interaction and uncertain
  maneuver prediction,'' \emph{{IEEE} Trans. on Intel. Veh.}, vol.~3, no.~1,
  pp. 5--17, 2018.

\bibitem{luo2018porca}
Y.~Luo, P.~Cai, A.~Bera, D.~Hsu, W.~S. Lee, and D.~Manocha, ``{PORCA}: Modeling
  and planning for autonomous driving among many pedestrians,'' \emph{{IEEE}
  Robot. Autom. Lett.}, vol.~3, no.~4, pp. 3418--3425, 2018.

\bibitem{cunningham2015mpdm}
A.~G. Cunningham, E.~Galceran, R.~M. Eustice, and E.~Olson, ``{MPDM}:
  Multipolicy decision-making in dynamic, uncertain environments for autonomous
  driving,'' in \emph{ICRA. IEEE}, 2015, pp. 1670--1677.

\bibitem{galceran2017multipolicy}
E.~Galceran, A.~G. Cunningham, R.~M. Eustice, and E.~Olson, ``Multipolicy
  decision-making for autonomous driving via changepoint-based behavior
  prediction: Theory and experiment,'' \emph{Auton. Robot.}, vol.~41, no.~6,
  pp. 1367--1382, 2017.

\bibitem{zhang2020efficient}
L.~Zhang, W.~Ding, J.~Chen, and S.~Shen, ``Efficient uncertainty-aware
  decision-making for automated driving using guided branching,'' in
  \emph{ICRA. IEEE}, 2020, pp. 3291--3297.

\bibitem{scokaert1998min}
P.~O. Scokaert and D.~Q. Mayne, ``Min-max feedback model predictive control for
  constrained linear systems,'' \emph{{IEEE} Trans. Autom. Ctrl.}, vol.~43,
  no.~8, pp. 1136--1142, 1998.

\bibitem{hardy2013contingency}
J.~Hardy and M.~Campbell, ``Contingency planning over probabilistic obstacle
  predictions for autonomous road vehicles,'' \emph{{IEEE} Trans. on Robot.},
  vol.~29, no.~4, pp. 913--929, 2013.

\bibitem{chen2022interactive}
Y.~Chen, U.~Rosolia, W.~Ubellacker, N.~Csomay-Shanklin, and A.~D. Ames,
  ``Interactive multi-modal motion planning with branch model predictive
  control,'' \emph{{IEEE} Robot. Autom. Lett.}, vol.~7, no.~2, pp. 5365--5372,
  2022.

\bibitem{da2022comprehensive}
F.~Da, ``Comprehensive reactive safety: No need for a trajectory if you have a
  strategy,'' in \emph{IROS. IEEE}, 2022, pp. 2903--2910.

\bibitem{Dosovitskiy17}
A.~Dosovitskiy, G.~Ros, F.~Codevilla, A.~Lopez, and V.~Koltun, ``{CARLA}: {An}
  open urban driving simulator,'' in \emph{Proceedings of the 1st Annual
  Conference on Robot Learning}, 2017, pp. 1--16.

\bibitem{ye2017despot}
N.~Ye, A.~Somani, D.~Hsu, and W.~S. Lee, ``{DESPOT}: Online {POMDP} planning
  with regularization,'' \emph{J. Artif. Intel. Res.}, vol.~58, pp. 231--266,
  2017.

\bibitem{kurniawati2016online}
H.~Kurniawati and V.~Yadav, ``An online {POMDP} solver for uncertainty planning
  in dynamic environment,'' in \emph{ISRR. Springer}, 2016, pp. 611--629.

\bibitem{ding2021epsilon}
W.~Ding, L.~Zhang, J.~Chen, and S.~Shen, ``{EPSILON}: An efficient planning
  system for automated vehicles in highly interactive environments,''
  \emph{{IEEE} Trans. on Robot.}, vol.~38, no.~2, pp. 1118--1138, 2021.

\bibitem{Chen2023TreestructuredPP}
Y.~Chen, P.~Karkus, B.~Ivanovic, X.~Weng, and M.~Pavone, ``Tree-structured
  policy planning with learned behavior models,'' \emph{ICRA. IEEE}, pp.
  7902--7908, 2023.

\bibitem{gonzalez2015review}
D.~Gonz{\'a}lez, J.~P{\'e}rez, V.~Milan{\'e}s, and F.~Nashashibi, ``A review of
  motion planning techniques for automated vehicles,'' \emph{{IEEE} Trans. on
  Intel. Trans. Syst.}, vol.~17, no.~4, pp. 1135--1145, 2015.

\bibitem{rufli2010design}
M.~Rufli and R.~Siegwart, ``On the design of deformable input-/state-lattice
  graphs,'' in \emph{ICRA. IEEE}, 2010, pp. 3071--3077.

\bibitem{mcnaughton2011motion}
M.~McNaughton, C.~Urmson, J.~M. Dolan, and J.-W. Lee, ``Motion planning for
  autonomous driving with a conformal spatiotemporal lattice,'' in \emph{ICRA.
  IEEE}, 2011, pp. 4889--4895.

\bibitem{ma2015efficient}
L.~Ma, J.~Xue, K.~Kawabata, J.~Zhu, C.~Ma \emph{et~al.}, ``Efficient
  sampling-based motion planning for on-road autonomous driving,'' \emph{{IEEE}
  Trans. on Intel. Trans. Syst.}, vol.~16, no.~4, pp. 1961--1976, 2015.

\bibitem{xu2012real}
W.~Xu, J.~Wei, J.~M. Dolan, H.~Zhao, and H.~Zha, ``A real-time motion planner
  with trajectory optimization for autonomous vehicles,'' in \emph{ICRA. IEEE},
  2012, pp. 2061--2067.

\bibitem{ziegler2014trajectory}
J.~Ziegler, P.~Bender, T.~Dang, and C.~Stiller, ``Trajectory planning for
  {B}ertha—a local, continuous method,'' in \emph{IV. IEEE}, 2014, pp.
  450--457.

\bibitem{chen2017constrained}
J.~Chen, W.~Zhan, and M.~Tomizuka, ``Constrained iterative lqr for on-road
  autonomous driving motion planning,'' in \emph{ITSC. IEEE}, 2017, pp. 1--7.

\bibitem{gu2015tunable}
T.~Gu, J.~Atwood, C.~Dong, J.~M. Dolan, and J.-W. Lee, ``Tunable and stable
  real-time trajectory planning for urban autonomous driving,'' in \emph{IROS.
  IEEE}, 2015, pp. 250--256.

\bibitem{fan2018baidu}
H.~Fan, F.~Zhu, C.~Liu, L.~Zhang, L.~Zhuang, D.~Li, W.~Zhu, J.~Hu, H.~Li, and
  Q.~Kong, ``Baidu {Apollo} {EM} motion planner,'' \emph{arXiv preprint
  arXiv:1807.08048}, 2018.

\bibitem{alsterda2019contingency}
J.~P. Alsterda, M.~Brown, and J.~C. Gerdes, ``Contingency model predictive
  control for automated vehicles,'' in \emph{ACC. IEEE}, 2019, pp. 717--722.

\bibitem{wang2022interaction}
R.~Wang, M.~Schuurmans, and P.~Patrinos, ``Interaction-aware model predictive
  control for autonomous driving,'' in \emph{ECC. IEEE}, 2023, pp. 1--6.

\bibitem{Kousik20}
S.~Kousik, S.~Vaskov, F.~Bu, M.~Johnson-Roberson, and R.~Vasudevan, ``Bridging
  the gap between safety and real-time performance in receding-horizon
  trajectory design for mobile robots,'' \emph{Intl. J. Robot. Res.}, vol.~39,
  08 2020.

\bibitem{Vaskov}
S.~Vaskov, U.~Sharma, S.~Kousik, M.~Johnson-Roberson, and R.~Vasudevan,
  ``Guaranteed safe reachability-based trajectory design for a high-fidelity
  model of an autonomous passenger vehicle,'' in \emph{ACC. IEEE}, 2019, pp.
  705--710.

\bibitem{rockafellar2002deviation}
R.~Rockafellar, S.~Uryasev, and M.~Zabarankin, ``Deviation measures in risk
  analysis and optimization,'' \emph{SSRN Electronic Journal}, 12 2002.

\bibitem{mayne1966second}
D.~Mayne, ``A second-order gradient method for determining optimal trajectories
  of non-linear discrete-time systems,'' \emph{Intl. J. Ctrl.}, vol.~3, no.~1,
  pp. 85--95, 1966.

\bibitem{zhao2019multi}
T.~Zhao, Y.~Xu, M.~Monfort, W.~Choi, C.~Baker, Y.~Zhao, Y.~Wang, and Y.~N. Wu,
  ``Multi-agent tensor fusion for contextual trajectory prediction,'' in
  \emph{CVPR}, 2019, pp. 12\,126--12\,134.

\bibitem{chang2019argoverse}
M.-F. Chang, J.~Lambert, P.~Sangkloy, J.~Singh, S.~Bak, A.~Hartnett, D.~Wang,
  P.~Carr, S.~Lucey, D.~Ramanan \emph{et~al.}, ``Argoverse: 3{D} tracking and
  forecasting with rich maps,'' in \emph{CVPR}, 2019, pp. 8748--8757.

\bibitem{kousik2017safe}
S.~Kousik, S.~Vaskov, M.~Johnson-Roberson, and R.~Vasudevan, ``Safe trajectory
  synthesis for autonomous driving in unforeseen environments,'' in \emph{DSCC.
  ASME}, vol. 58271, 2017.

\bibitem{hwang2003applications}
I.~Hwang, D.~M. Stipanovic, and C.~J. Tomlin, ``Applications of polytopic
  approximations of reachable sets to linear dynamic games and a class of
  nonlinear systems,'' in \emph{ACC. IEEE}, vol.~6, 2003, pp. 4613--4619.

\bibitem{backupCBF}
Y.~Chen, M.~Jankovic, M.~Santillo, and A.~D. Ames, ``Backup control barrier
  functions: Formulation and comparative study,'' in \emph{CDC. IEEE}, 2021,
  pp. 6835--6841.

\bibitem{failsafe}
C.~Pek and M.~Althoff, ``Fail-safe motion planning for online verification of
  autonomous vehicles using convex optimization,'' \emph{{IEEE} Trans. on
  Robot.}, vol.~PP, 12 2020.

\bibitem{inpass}
M.~Elbanhawi, M.~Simic, and R.~Jazar, ``In the passenger seat: Investigating
  ride comfort measures in autonomous cars,'' \emph{ITSM. IEEE}, vol.~7, no.~3,
  pp. 4--17, 2015.

\bibitem{glaser2010maneuver}
S.~Glaser, B.~Vanholme, S.~Mammar, D.~Gruyer, and L.~Nouveliere,
  ``Maneuver-based trajectory planning for highly autonomous vehicles on real
  road with traffic and driver interaction,'' \emph{{IEEE} Trans. on Intel.
  Trans. Syst.}, vol.~11, no.~3, pp. 589--606, 2010.

\end{thebibliography}

\end{document}